\documentclass[11pt,a4paper]{article}
\usepackage[hyperref]{emnlp2020}
\usepackage{times}
\usepackage{latexsym}

\usepackage{color}
\usepackage{comment}
\usepackage{bm}
\usepackage{setspace}
\usepackage{multirow}
\usepackage{graphicx}
\usepackage{amsfonts}
\usepackage{amsmath}
\usepackage{amsthm}

\usepackage{times}
\usepackage{soul}
\usepackage{url}
\usepackage{paralist}
\usepackage{amssymb}

\usepackage{microtype}

\aclfinalcopy 

\title{Variational Gaussian Topic Model with Invertible Neural Projections}

\author{Rui Wang$^{\S,\dag,\ddag,*}$\ \ \ \  Deyu Zhou$^{\dag,}\thanks{co-corresponding author.}$\ \ \ \ \textbf{Yuxuan Xiong}$^{\dag}$\ \ \ \ \textbf{Haiping Huang}$^{\S,\ddag}$ \\
     $^\S$School of Computer Science, Nanjing University of Posts and Telecommunications, China \\
	$^{\dag}$School of Computer Science and Engineering, Key Laboratory of Computer Network\\
	and Information Integration, Ministry of Education, Southeast University, China \\
     $^{\ddag}$Jiangsu High Technology Research Key Laboratory for Wireless Sensor Networks, Nanjing, China.\\ 
		rui\_wang@njupt.edu.cn,\ \{d.zhou, yuxuanxiong\}@seu.edu.cn,\ hhp@njupt.edu.cn} 

\date{}

\begin{document}
\maketitle
\begin{abstract}
Neural topic models have triggered a surge of interest in extracting topics from text automatically since they avoid the sophisticated derivations in conventional topic models. However, scarce neural topic models incorporate the word relatedness information captured in word embedding into the modeling process. To address this issue, we propose a novel topic modeling approach, called Variational Gaussian Topic Model (VaGTM). Based on the variational auto-encoder, the proposed VaGTM models each topic with a multivariate Gaussian in decoder to incorporate word relatedness. Furthermore, to address the limitation that pre-trained word embeddings of topic-associated words do not follow a multivariate Gaussian, Variational Gaussian Topic Model with Invertible neural Projections (VaGTM-IP) is extended from VaGTM. Three benchmark text corpora are used in experiments to verify the effectiveness of VaGTM and VaGTM-IP. The experimental results show that VaGTM and VaGTM-IP outperform several competitive baselines and obtain more coherent topics.
\end{abstract}

\section{Introduction}
Topic models have been extensively explored in the natural language processing (NLP) community for unsupervised knowledge discovery. Many variants~\cite{lin2009joint,zhou2014simple} of Latent Dirichlet Allocation (LDA)~\cite{blei2003latent} have been proposed to tackle different extraction tasks. And a large body of work has considered approximate inference methods which need sophisticated mathematical derivations for model inference.

To solve this limitation, developing the neural-based topic models which employ black-box inference mechanism with neural network seems to be a promising direction. Based on variational auto-encoder (VAE)~\cite{kingma2013auto}, Miao and Yu~\shortcite{miao2016neural} propose the Neural Variational Document Model (NVDM) which uses a decoder to reconstruct the document by generating the words independently. However, the inappropriate Gaussian prior employed in NVDM may result in bad topic quality. Thus, Srivastava and Sutton~\shortcite{srivastava2017autoencoding} propose NVLDA and ProdLDA, neural topic models based on VAE, in which the logistic normal distribution is used as the prior for topic extraction. Besides, Wang~\shortcite{wang2019atm} proposes the Adversarial-neural Topic Model (ATM) based on the adversarial training.


On the other hand, pre-trained word embeddings (e.g. \emph{GloVe}~\cite{pennington2014glove} and \emph{word2vec}~\cite{mikolov2013efficient}) from massive unlabeled text corpora provide a way of injecting word relatedness into models that would otherwise treat words as isolated categories. Thanks to the rich lexico-semantic regularities in language contained in word embedding, the state-of-the-art performance of various NLP task (e.g. sentiment analysis~\cite{majumder-etal-2018-iarm}, stance detection~\cite{sun-etal-2018-stance}) have been significantly improved through incorporating word embeddings. However, scarce similar attempts have been made in neural topic modeling. 

Thus, in this paper, we propose a Variational Gaussian Topic Model (VaGTM) which incorporates word embeddings into topic modeling process. Based on the variational auto-encoding, {\color{black} its principle idea is to build a decoder which is able to reconstruct the observed documents using pre-trained word embeddings along with inferred topic distributions.} Unlike the neural-based approaches that only use word co-occurrence information, VaGTM models each topic with a multivariate Gaussian distribution in decoder, and the probability of a word in a specific topic could be calculated by topic-associated Gaussian probability density function using word embeddings as input. Instead of providing an analytic approximation, the VaGTM employs the variational lower bound of the observed documents as training objective to learn the means and covariance matrix of topic-associated Gaussian distributions. Due to the semantic properties of word embeddings, these distributions in decoder could capture the underlying thematic structures of text collections. 

Moreover,  to address the limitation that {\color{black}pre-trained word representations of topic-associated words may not follow a multivariate Gaussian since the specific properties of language captured by any word embedding scheme is difficult to control, we employ a flow-based transformation network~\cite{dinh2014nice} to project the original word embedding space into a {\color{black}mixed Gaussian embedding space} and propose the Variational Gaussian Topic Model with Invertible neural Projections (VaGTM-IP).
}

Our contributions are summarized below:
\begin{itemize}
\item We propose a novel Variational Gaussian Topic Model (VaGTM) which could incorporate the semantic relatedness in word embeddings into topic modeling process.
\item {\color{black}To deal with the limitation that word embeddings of topic-associated words do not follow a multivariate Gaussian distribution, we extend the VaGTM and propose the Variational Gaussian Topic Model with Invertible neural Projections (VaGTM-IP).}
\item Experimental results on three public datasets show that VaGTM and VaGTM-IP outperform the state-of-the-art approaches in terms of four topic coherence measures.
\end{itemize}

\section{Related Work}
{\color{black}Our work is related to three lines of research, word representation learning, invertible projection learning and neural topic modeling.

\subsection{Word Representation Learning}
Distributed semantic models (i.e. word embeddings) have recently been applied successfully in many NLP tasks.

Neural based models, such as \emph{word2vec}, have been more efficient thanks to the skip-gram with a negative sampling training method~\cite{mikolov2013efficient}. To solve the limitation that \emph{word2vec} only employs local context information, Pennington~\shortcite{pennington2014glove} proposed \emph{GloVe}, a global log-bilinear regression model, which combines the advantages of the global matrix factorization and local context window methods. To generate a vector for an out of vocabulary word, \emph{FastText}~\cite{joulin2016bag,athiwaratkun2018probabilistic}, has been proposed. It treats each word as made of character n-grams and word vectors are then computed from the sum of their n-gram representations. 

\begin{figure*}[!h]
\centering
\includegraphics[
  width=0.85\textwidth,
  keepaspectratio]
{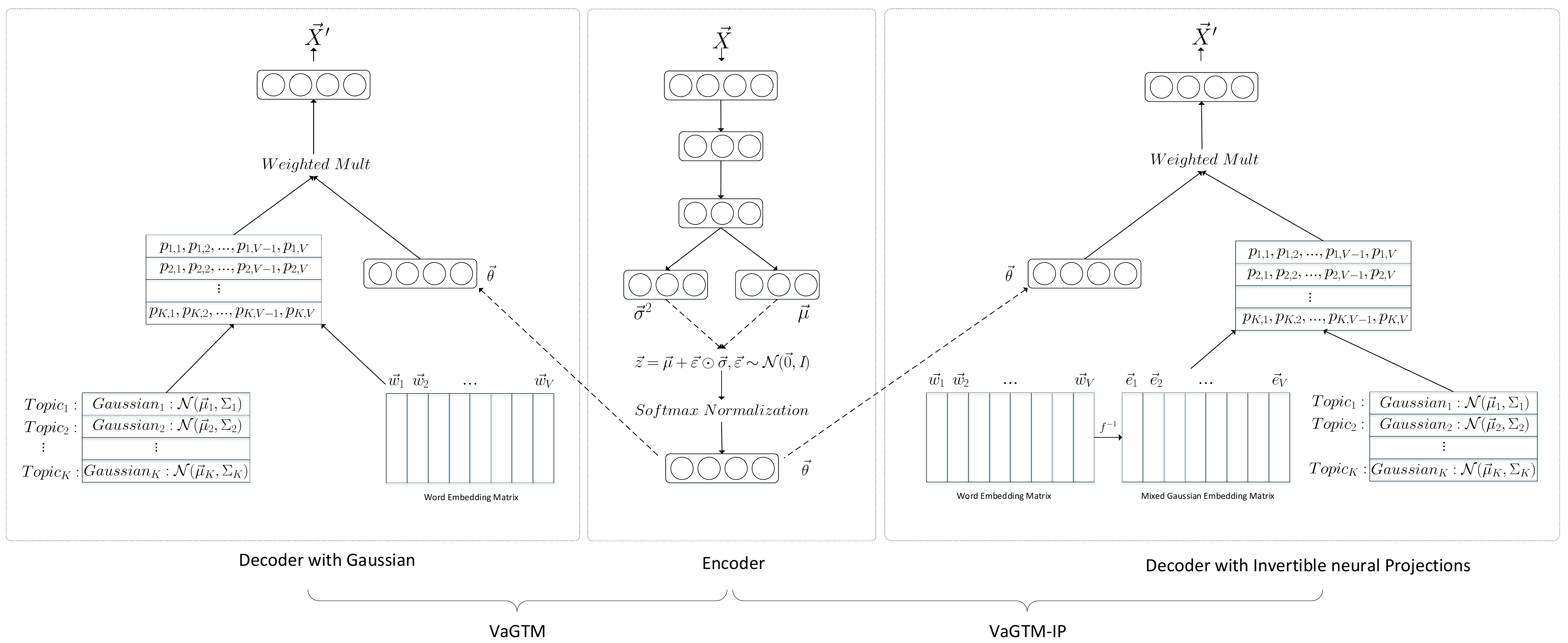}
\caption{The framework of the Variational Gaussian Topic Model (VaGTM) and Variational Gaussian Topic Model with Invertible neural Projections (VaGTM-IP).}

\label{fig:vae_framework}
\end{figure*}

\subsection{Invertible Projection Learning}
Invertible Projection Learning, also known as generative flows, is first described in NICE~\cite{dinh2014nice} and has recently received much attention~\cite{dinh2016density,jacobsen2018revnet,kingma2018glow}. 

Typically, generative flows have been proposed for image generation. NICE~\cite{dinh2014nice} firstly designed a composition of additive coupling layers to non-linearly project a complex high-dimensional densities into a simple prior and the inverse projection is used for generating images. 
To address the issue that variational auto-encoder could not approximate complex posterior, Danilo~\shortcite{Rezende:2015:VIN:3045118.3045281} incorporated the normalizing flow into variational inference for specifying arbitrarily complex posterior distributions. Also, He~\cite{he-etal-2018-unsupervised} incorporated the generative flow into the Hidden Markov Model for unsupervised part-of-speech tagging.


\subsection{Neural Topic Modeling}
To overcome the difficult exact inference of topic models based on the directed graph, a replicated softmax model, called RSM, based on the Restricted Boltzmann Machines was proposed in ~\cite{hinton2009replicated}. Inspired by the variational auto-encoder, Miao et al.~\shortcite{miao2016neural} used the multivariate Gaussian as the prior distribution of latent space and proposed the Neural Variational Document Model (NVDM) for text modeling. Recently, to deal with the inappropriate Gaussian prior of NVDM, Srivastava and Sutton~\shortcite{srivastava2017autoencoding} proposed the NVLDA and ProdLDA which approximates the Dirichlet prior using a logistic normal distribution. Furthermore, Dieng~\shortcite{dieng2019topic} proposed the ETM based on VAE which models topics in embedding space. 


{\color{black}Although the extensive exploration of the above fields, scarce work has been done to incorporate word representation learning and invertible projection learning into neural topic modeling. Thus, we propose two novel topic modeling approaches, called VaGTM and VaGTM-IP, which differs published works in the following facets: (1). Unlike the NVDM, NVLDA, ProdLDA and ATM which only uses word co-occurrence information, our proposed VaGTM models each topic with a multivariate Gaussian distribution which could incorporate the word semantic relatedness into the modeling process; (2). {\color{black}To deal with the issue that word embeddings of topic-associated words do not follow a multivariate Gaussian, we  incorporate a flow-based projector into the modeling process and propose the VaGTM-IP }. }}

\section{Methodologies}
Based on the encoder-decoder framework,  we propose the Variational Gaussian Topic Model (VaGTM) and the Variational Gaussian Topic Model with Invertible neural Projections (VaGTM-IP).

As shown in {\color{black}Fig.~\ref{fig:vae_framework}}, VaGTM is constituted by an encoder and a decoder with Gaussian, and VaGTM-IP contains an encoder and a decoder with invertible neural projections. The encoder, shared by VaGTM and VaGTM-IP, uses the bag of word representations with \emph{tf-idf} weights $\vec X\in \mathbb{R}^{V}$ as input to approximate the intractable posterior distribution over the latent topics $\vec \theta\in \mathbb{R}^{K}$, and decoders (in both models) aim to reconstruct $\vec X$ from the encoded topic distribution $\vec \theta$ and pre-trained word embeddings. In VaGTM, decoder incorporates the word relatedness by modeling topics with multivariate Gaussian. To further address the limitation that word embeddings of topic-associated words do not follow a multivariate Gaussian, an invertible neural projection is used in the decoder of VaGTM-IP. We explain the design of these networks in more detail below.

\subsection{Encoder network}

 To obtain interpretable topics, Dirichlet prior over topics  is commonly used~\cite{wallach2009rethinking}.
 However, it is difficult to explicitly model topics with Dirichlet prior since it is hard to develop an effective  reparameterization. Thus, we solve the issue by constructing a Laplace approximation~\cite{hennig2012kernel}.


 Aiming at compressing the weighted bag of words representation $\vec X$ into the latent topic distribution $\vec \theta$, the encoder $q_{\bm{\psi}}(\vec \theta|\vec X)$, parameterized with $\bm{\psi}$, contains five layers  which are one $V$-dimensional document representation layer, two $H$-dimensional semantic extraction layers, one {\color{black}$K$-dimensional reparameterization layer} and one $K$-dimensional topic distribution layer as shown in {\color{black} Fig.~\ref{fig:vae_framework}}. Firstly, it projects $\vec X$ into an $H$-dimensional semantic space through semantic extraction layers based on the transformation:
\begin{align}
\vec a_{s}^{1}&= \ln (1+\exp (W_{s}^{1}\vec X+\vec b_{s}^{1}))\\
\vec a_{s}^{2}&=\ln (1+\exp(W_{s}^{2}\vec a_{s}^{1}+\vec b_{s}^{2}))
\end{align}
where $W_{s}^{1}\in \mathbb{R}^{H \times V}$ and $W_{s}^{2}\in \mathbb{R}^{H \times H}$ are weight matrices, $\vec b_{s}^{1}$ and $\vec b_{s}^{2}$ are corresponding basis terms, $\vec a_{s}^{1}$ and $\vec a_{s}^{2}$ are the semantic representations of $\vec X$, $\exp(\cdot)$ is the element-wise exponential function. 

To further infer the posterior distribution over topics, following the Laplace approximation, a Gaussian distribution should be generated. Thus, the encoder projects $\vec a_{s}^{2}$  into two $K$-dimensional Gaussian parameters $\vec \mu$ and $\vec{\sigma^{2}} $ using:
 \begin{align}
\vec \mu&= \textrm{BN}(W_{r}^{\mu}\vec a_{s}^{2}+\vec b_{\mu})
\\ \vec{\sigma^{2}}&=\exp(\textrm{BN}(W_{r}^{\sigma}\vec a_{s}^{2}+\vec b_{\sigma}))
\end{align}
where $W_{r}^{\mu}\in \mathbb{R}^{K\times H}$, $W_{r}^{\sigma}\in \mathbb{R}^{K \times H}$, $\vec b_{\mu}$ and $\vec  b_{\sigma}$ are weight matrices and basis terms of reparameterization layers,   $\textrm{BN}(\cdot)$ is batch normalization. And $\vec \mu$ and $\vec{\sigma^{2}}$ are mean and diagonal covariance of posterior Gaussian corresponding to input $\vec X$. 

Finally, based on the reparameterization trick~\cite{kingma2013auto} and Laplace approximation~\cite{hennig2012kernel},  $\vec X$ could be inferred to a $K$-dimensional topic distribution $\vec \theta$ using below transformation:
 \begin{align}
\vec z&=\vec \mu+\vec \varepsilon\odot \vec \sigma, \vec \varepsilon\sim \mathcal N(\vec 0,\emph{I}) \label{sample}\\
\vec \theta&=\text{softmax}(\vec z) \label{softmax}
\end{align}
Here, $\mathcal N(\vec 0,\emph{I})$ is $K$-dimensional standard multivariate Gaussian, and the posterior distribution $q(\vec \theta|\vec X)$ follows the Dirichlet distribution in softmax basis~\cite{mackay1998choice}. 

\subsection{Decoder with Gaussian}
The decoder $p_{\bm{\omega}}(\vec X'|\vec \theta)$, parameterized with $\bm{\omega}$, reconstructs the documents by independently generating the words ($\vec \theta\rightarrow \vec X'_{i}$). Besides, to incorporate the word relatedness captured in word embeddings, VaGTM models each topic with a multivariate Gaussian as depicted in the left panel of Fig.~\ref{fig:vae_framework}.

Concretely,  VaGTM employs multivariate Gaussian $\mathcal N(\vec \mu_{k},\Sigma_{k})$ to model the $k$-th topic. where  $\vec \mu_{k}$ and $\Sigma_{k}$ represent mean and covariance matrix. Following the probability density of Gaussian, for each word $v\in \{1, 2,...,V\}$, its probability in the $k$-th topic $\phi_{k,v}$ could be calculated as:
\begin{align}
p(\vec w_{v}|&topic=k)=\mathcal N(\vec w_{v};\vec \mu_{k}, \Sigma_{k}) \notag \\
=&\frac{\exp(-\frac{1}{2}(\vec w_{v}-\vec \mu_{k})^{\text{T}}\Sigma^{-1}(\vec w_{v}-\vec \mu_{k}))}{\sqrt{(2\pi)^{D_{w}}|\Sigma_{k}|} }\label{distribution_1} \\
\phi_{k,v}=&\frac{p(\vec w_{v}|topic=k)}{\sum_{v=1}^{V}p(\vec w_{v}|topic=k)}\label{distribution_2}
\end{align}
where $\vec w_{v}$ is the word embedding of word $v$,  $V$ is the vocabulary size, $|\Sigma_{k}|=\rm det$ $\Sigma_{k}$ represents the determinant of covariance matrix $\Sigma_{k}$, $D_{w}$ is the dimension of word embeddings, $\vec \phi_{k}$ is the normalized word distribution of the $k$-th topic.

To reconstruct the document $\vec X$, topic-word distributions and the encoded topic distribution $\vec \theta$ are combined using:
\begin{align}
p_{rec}(\vec v|\vec \theta)= \sum_{k=1}^{K}\vec \phi_{k}\cdot \theta_{k} \label{reconstruct_distribution2}
\end{align} 
where $K$ denotes the topic number, $\theta_{k}$ means the proportion of $k$-th topic in $\vec X$,  and $p_{rec}(v|\vec \theta)$ is the conditional distribution over the vocabulary which is employed to reconstruct the $\vec X$ by maximizing the equation:
\begin{equation}
p(\vec X'|\vec \theta)=\sum_{v=1}^{V}X_{v}\cdot\log \left[ p_{rec}(v|\vec \theta)\right]
\end{equation}
{\color{black}where $X_{v}$ is the \emph{tf-idf} weight of the $v$-th word in document $\vec X$, and $\vec X'$ is the reconstructed document.} 

\subsection{Decoder with Invertible neural Projections}
Word embeddings indeed encode numerous semantic regularities. However, widely used word representation scheme, such as \emph{word2vec}~\shortcite{mikolov2013efficient} and \emph{Glove}~\shortcite{pennington2014glove}, do not model embeddings with Gaussian and the resulting embeddings often follow a complicated distribution $\mathbb{P}_{w}$. {\color{black}Thus, it is not accurate to approximate $\mathbb{P}_{w}$ with mixed Gaussian $\mathbb{P}_{m}$ as VaGTM. 

To approximate the $\mathbb{P}_{w}$ and yield a mixed Gaussian embedding space which is more suitable for topic modeling, a non-linear neural projector $f(\cdot)$ is employed to deterministically transform the mixed Gaussian embedding space to the pre-trained word embedding space. The vector representation of word $v$ in mixed Gaussian embedding space is denoted as $\vec e_{v}\in \mathbb{R}^{D_{e}}$, $D_{e}$ is the dimension of mixed Gaussian embedding space. Thus, for each word $v$, it has the transformation below: 
\begin{equation}
\vec w_{v}=f(\vec e_{v})
\end{equation}
\ \ \ \ Due to the reconstruction of $\vec X$ is conditioned on word embeddings, as shown in the right panel of Fig.~\ref{fig:vae_framework}, the projector $f(\cdot)$ should be invertible. To this end, a flow-based projector~\cite{dinh2014nice} is employed here to bridge the mixed Gaussian embedding space and word embedding space since it can transform a simple distribution into a complicated one. 

Thus, decoding with invertible neural projection, the probability of word $v$ in the $k$-th topic is proportional to $p(\vec w_{v}|topic=k)$ which is defined as:
\begin{align}
p(&\vec w_{v}|topic=k)=\int p(\vec w_{v}|\vec e_{v})\cdot p(\vec e_{v}|topic =k)d\vec e_{v}\notag \\
&=\int \square \cdot \mathcal N(f^{-1}(\vec w_{v});\vec \mu_{k},\Sigma_{k})\left|\mathrm{det}\frac{\partial f^{-1}}{\partial \vec w_{v}}\right|d\vec w_{v} \notag\\
&=\mathcal N(f^{-1}(\vec w_{v});\vec \mu_{k}, \Sigma_{k})\cdot \left|\mathrm{det}\frac{\partial f^{-1}}{\partial \vec w_{v}} \right| \label{vagtm_dis} 
\end{align}
where $\frac{\partial f^{-1}}{\partial \vec w_{v}}$ represents the Jacobian matrix of inverse projection function $f^{-1}(\cdot)$ at $\vec w_{v}$ which is nonzero and differentiable if and only if $f^{-1}(\cdot)$ exists, and $\left|\mathrm{det}\frac{\partial f^{-1}}{\partial \vec w_{v}} \right|$ denotes the absolute value of its determinant. The symbol $\square$ in Eq.~\ref{vagtm_dis} represents $\delta (\vec w_{v}-f(\vec e_{v}))$. Here, $\delta(\cdot)$ is the Dirac delta function centered at $f(\vec e_{v})$ which is defined as:
\begin{equation}
p(\vec w_{v}|\vec e_{v})=\delta(\vec w_{v}-f(\vec e_{v}))=\left\{ 
\begin{aligned}
\infty & & \vec w_{v}=f(\vec e_{v})\\
0 & & \mathrm{otherwise}
\end{aligned}
\right.
\end{equation}

Comparing Eq.~\ref{distribution_1} and Eq.~\ref{vagtm_dis}, it should be observed that the addition of the Jacobian term will increase the difficulty of optimizing as it requires that all component functions be invertible and also requires storage of large Jacobian matrices. {\color{black}To address this issue, we use additive coupling layer suggested in~\cite{dinh2014nice} to guarantee a unit Jacobian determinant and the invertibility.} To reconstruct the document $\vec X$, it could be observed from Eq.~\ref{vagtm_dis} that only $f^{-1}(\cdot)$ is required. Thus, we explicitly design the inverse projector $f^{-1}(\cdot)$ as Fig.~\ref{fig:flow}.
\begin{figure}[!ht]
\centering
\includegraphics[
  width=0.42\textwidth,
  keepaspectratio]
{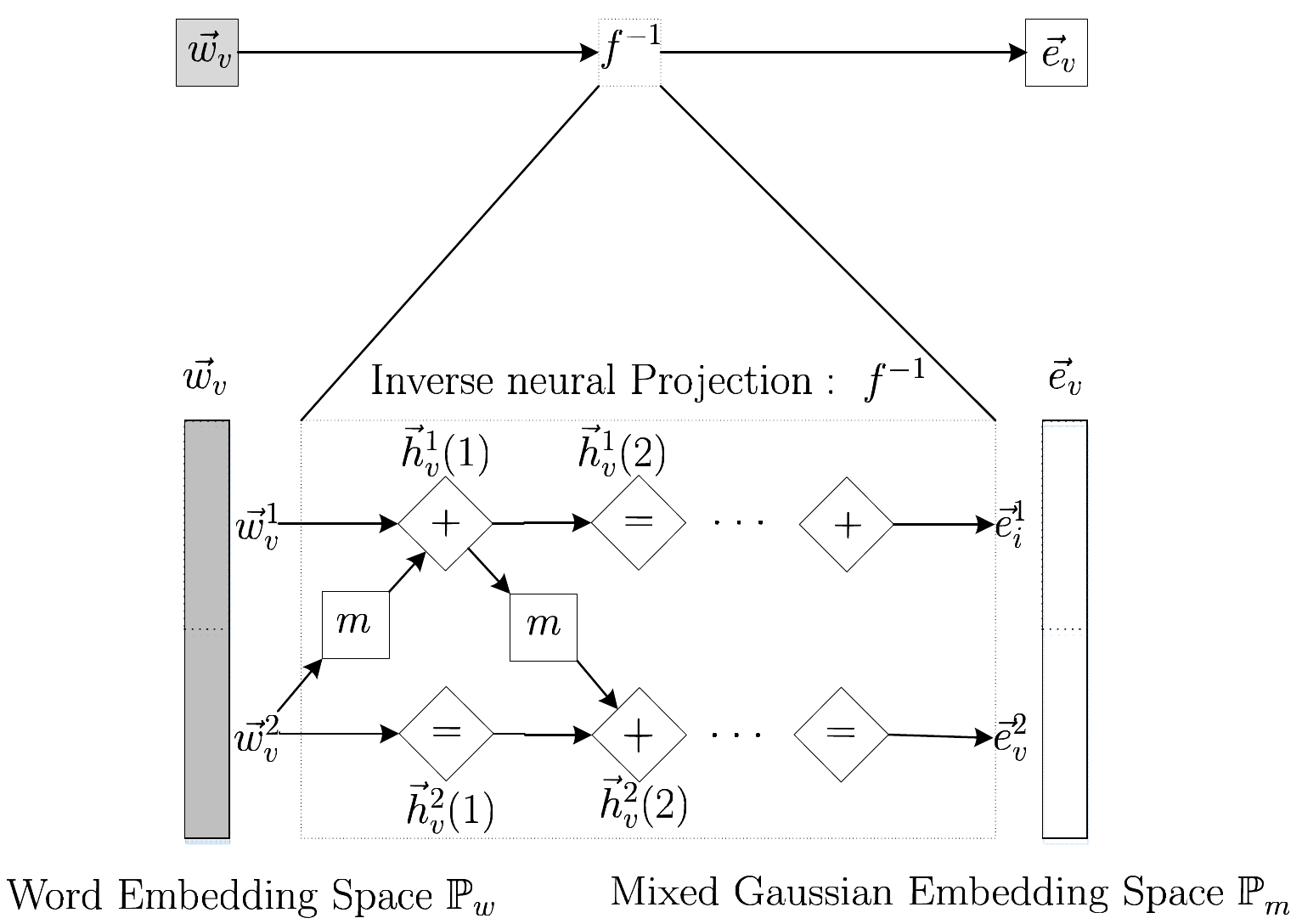}
\caption{Description of the invertible neural projection $f^{-1}(\cdot)$, it transforms the pre-trained word embedding $\vec w_{v}$ to a point $\vec e_{v}$ in {\color{black}mixed Gaussian embedding space}. }
\label{fig:flow}
\end{figure}


To accomplish the transformation,  word embedding $\vec w_{v}$ is first partitioned into two halves of dimensions, $\vec w^{1}_{v}$ and $\vec w^{2}_{v}$, respectively. And the first coupling layer is represented by the nonlinear transformation from $\vec w_{v}$ to $\vec h_{v}(1)$ which could be defined as:
\begin{align}
\vec h_{v}^{1}(1)=\vec w_{v}^{1}, \quad 
\vec h_{v}^{2}(1)=\vec w_{v}^{2}+m(\vec w_{v}^{1}) 
\end{align}
where $m(\cdot):\mathbb{R}^{D_{w}/2}\rightarrow \mathbb{R}^{D_{w}/2}$  is the \emph{coupling function} which could be designed as any nonlinear function with same input and output shape. We choose  additive coupling layer to reduce the computational consumption. To build a more complex transformation, we compose several \emph{coupling layers} and exchange the role of two half vectors at each  layer to ensure that the composition of two layers modifies every dimension. 

Then, following the reconstruction procedure illustrated in decoding process of VaGTM (subsection 3.2), the document $\vec X$ could be reconstructed in a similar way in VaGTM-IP. 

\subsection{Variational Objective}

In variational Bayesian, the variational lower bound on the marginal likelihood of document $\vec X$ is usually formed as:
\begin{align}
&\mathcal{L}(\vec X;\bm{\psi}, \bm{\omega})=\notag\\ &\mathbb{E}_{q_{\bm{\psi}(\vec \theta|\vec X)}} \log p_{\bm{\omega}}(\vec X|\vec \theta)  -KL(q_{\bm{\psi}}(\vec \theta|\vec X)\|p_{\omega}(\vec \theta)) \label{elbo} 
\end{align}
where $\vec \theta$ denotes the topic distribution of $\vec X$. To mimic the Dirichlet prior over $\vec \theta$,  the Laplace approximation is used. and the variational lower bound is rewrited as:
\begin{align}
\mathcal{L}(\vec X;\bm{\psi}, \bm{\omega}&)=\sum_{v=1}^{V}X_{v}\cdot\log \left[ p_{rec}(v|\vec \theta)\right] \notag\\ -\frac{1}{2}&\left[\mathrm{tr(\Sigma^{-1}\Sigma_{0})}+\Delta-K+\ln \frac{\mathrm{det}\ \Sigma}{\mathrm{det}\ \Sigma_{0}} \right]
\end{align}
where $\Delta=(\vec \mu-\vec \mu_{0})^{\mathsf{T}} \Sigma^{-1}(\vec \mu-\vec \mu_{0})$, $\vec \mu_{0}$ and $\Sigma_{0}$ are means and covariance matrix of standard multivariate Gaussian,  and $p_{rec}(v|\vec\theta)$ is the $v$-th dimension of reconstruction distribution defined as Eq.~\ref{reconstruct_distribution2}.

  \begin{figure*}[!ht]
\centering
\includegraphics[
  width=0.8\textwidth,
  keepaspectratio]
{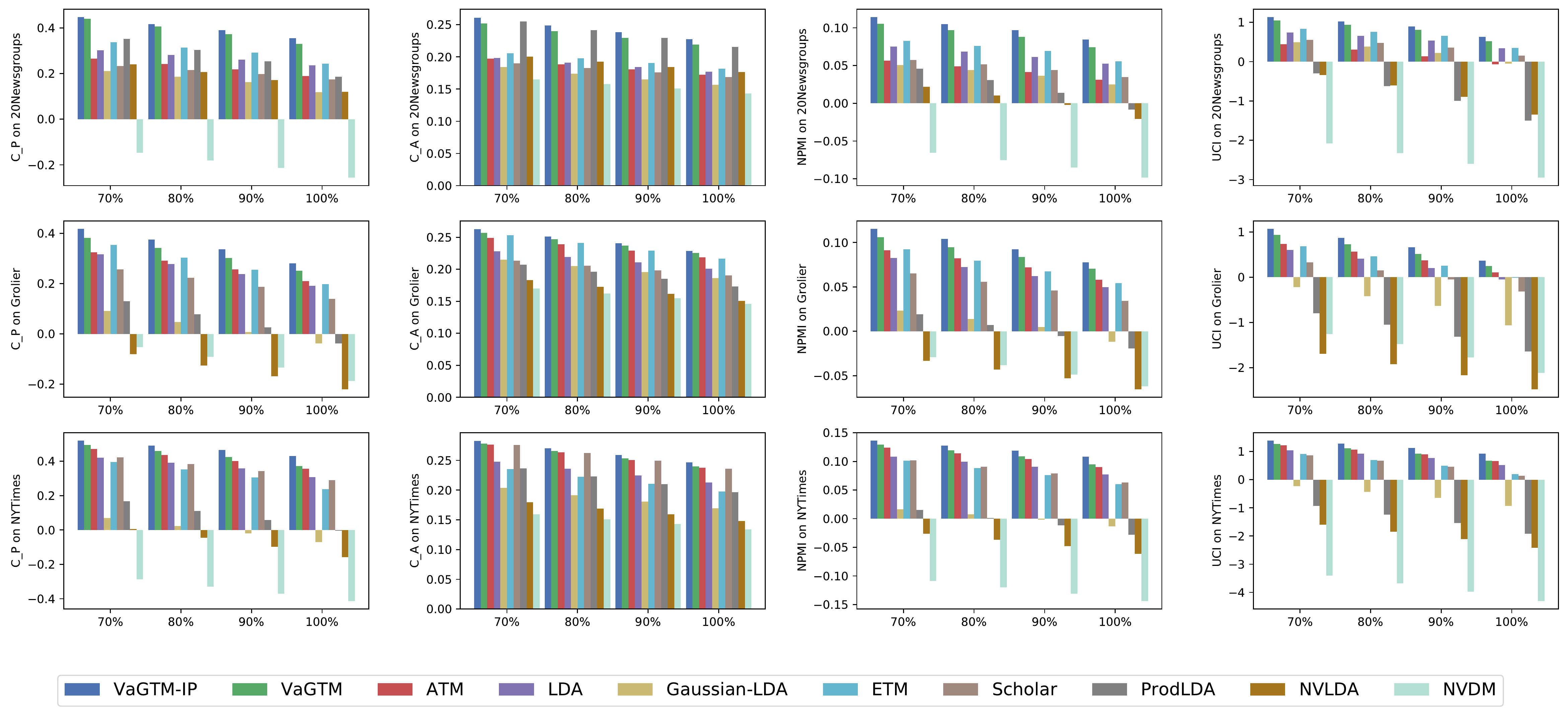}
\caption{The comparison of average topic coherence vs. different topic proportion on 20Newsgroups, Grolier and NYTimes.}
\label{fig:cmp_bar}
\end{figure*}

\section{Experiments}
In this section, we first introduce the text corpora and then describe the compared baselines. Finally, we present the experimental results.

\subsection{Experiment Settings}
To validate the effectiveness of proposed VaGTM and VaGTM-IP for topic modeling, 20Newsgroups~\footnote{http://qwone.com/ jason/20Newsgroups/} dataset,  Grolier~\footnote{https://cs.nyu.edu/~roweis/data/} dataset and NYTimes~\footnote{http://archive.ics.uci.edu/ml/datasets/Bag+of+Words} dataset are selected. Details are summarized below: 

\begin{compactitem}
\item \emph{\underline{20Newsgroups dataset}}, provided in~\cite{lang1995newsweeder}, it is a collection of approximately 20,000 newsgroup articles, partitioned evenly across 20 different newsgroups. 
\item \emph{\underline{Grolier dataset}} is built from Grolier Multimedia Encycopedia, and its content covers almost all the fields in the world, such as religious, technology, economics and etc.
\item \emph{\underline{NYTimes dataset}} is a collection of news articles published between 1987 and 2007, and the dataset has a wide range of topics, such as sports, politics and etc. 
\end{compactitem}

Besides, we choose the following approaches as baselines:
\begin{compactitem}
\item \textbf{\underline{LDA}}~\cite{blei2003latent}, is a topic model that generates topics based on bag-of-words assumption. We implement the LDA model with the suggested configuration~\cite{griffiths2004finding}.
\item \textbf{\underline{Gaussian-LDA}}~\cite{das2015gaussian}, is a conventional topic model  which uses the word embeddings information. The original implementation is used in this paper~\footnote{https://github.com/rajarshd/Gaussian LDA}. 
\item \textbf{\underline{NVDM}}~\cite{miao2016neural} is an unsupervised text modeling approach based on variational auto-encoder. We use the original implementation in the paper~\footnote{https://github.com/ysmiao/nvdm}.
\item \textbf{\underline{NVLDA}}~\cite{srivastava2017autoencoding}, is a neural topic model based on VAE. It models topics with logistic normal prior, we use the original implementation~\footnote{https://github.com/akashgit/autoencoding vi for topic models}.
\item \textbf{\underline{ProdLDA}}~\cite{srivastava2017autoencoding}, is a variant of NVLDA, in which the distribution over individual words is a product of experts rather than the mixture model. 
\item \textbf{\underline{Scholar}}~\cite{card2018neural}, is a neural topic modeling approach based on variational autoencoder, we use the original implementation~\footnote{https://github.com/dallascard/scholar}. 
\item \textbf{\underline{ETM}}~\cite{dieng2019topic}, is a neural topic model on embedding space, we use the original implementation\footnote{https://github.com/adjidieng/ETM}. 
\item \textbf{\underline{ATM}}~\cite{wang2019atm}, is a neural topic modeling approach based on adversarial training, we re-implement the ATM follow the default parameter settings.
\end{compactitem}

\begin{figure*}[!h]
\centering
\includegraphics[
width=0.8\textwidth,
  keepaspectratio
  ]
{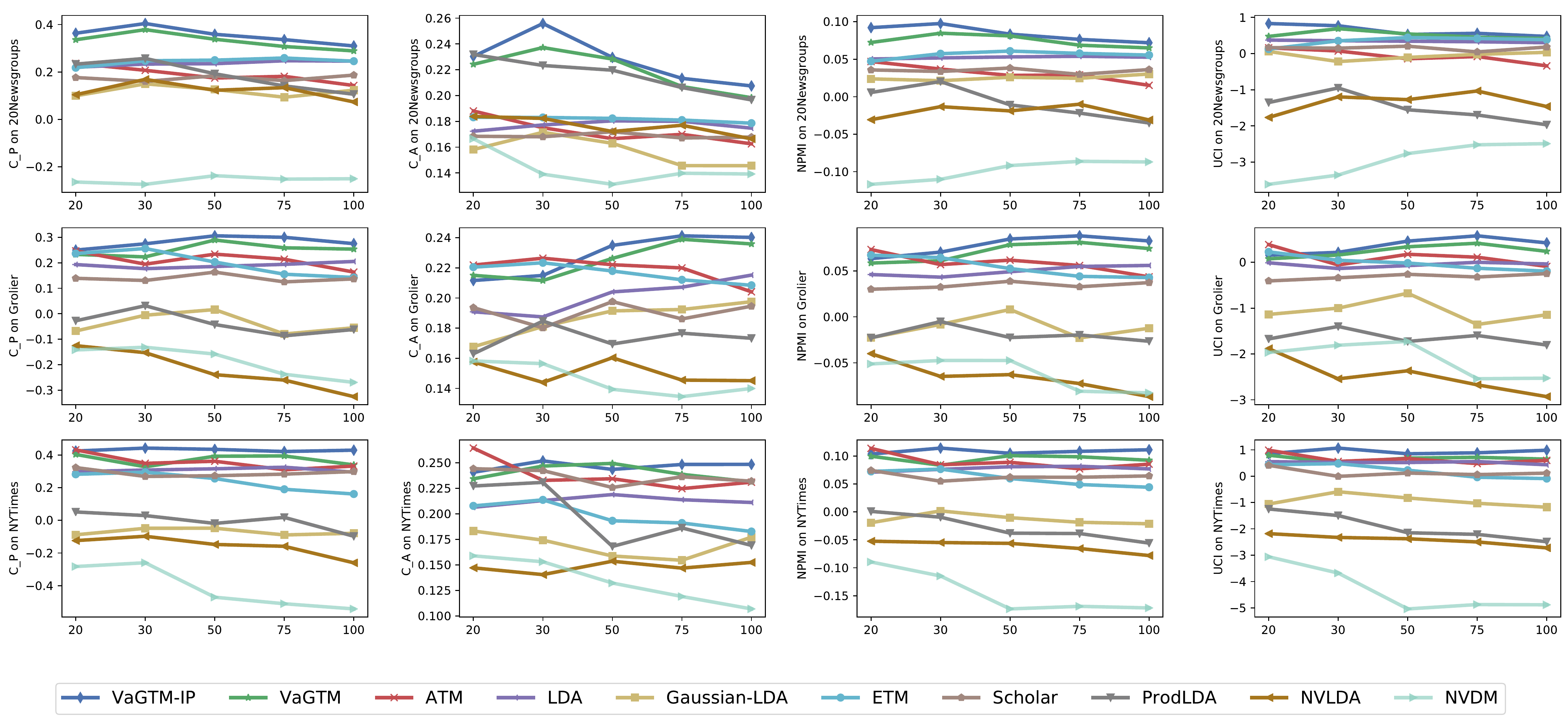}
\caption{The comparison of average topic coherence vs. different topic number on 20Newsgroups, Grolier and NYTimes.}
\label{cmpcurve}
\end{figure*}

\begin{table}[!ht]
\centering
\footnotesize
\scalebox{0.9}{
\begin{tabular}{l|c|c}
\hline 
\textbf{Dataset}& \textbf{\#Documents}&\textbf{ \# Words} \\
\hline 
20Newsgroups&11,259&1,995\\
Grolier&29,762&15,276\\
NYTimes&99,992&12,604\\
\hline
\end{tabular}}
\caption{The statistics of the processed datasets}
\label{tb:statistics}
\end{table}
For the NYTimes dataset, 100,000 articles are randomly selected, and low frequency words are removed here. The statistics of the processed datasets are listed in Table~\ref{tb:statistics}.

\subsection{Topic Coherence Evaluation}

Typically, the likelihood of held-out documents and topic coherence metrics are used to evaluate topic modeling performance. However, as pointed out in~\cite{NIPS2009_3700} the likelihood of held-out documents doesn't correspond to human judgment, we follow~\cite{Roder:2015:EST:2684822.2685324} and choose {\color{black}four} topic coherence metrics which are C\_P, C\_A, NPMI and UCI to evaluate the extracted topics, and higher value implies more coherent topic. All the topic coherences are computed with the Palmetto\footnote{https://github.com/dice-group/Palmetto} library and each topic is represented by top ten words according to the topic-word distribution.

\begin{table}[!t]
\centering
\footnotesize
\scalebox{0.78}{
\begin{tabular}{c|l|ccccc}
\hline
{\bfseries Dataset}&{\bfseries Model}& {\bfseries C\_P}&{\bfseries C\_A}&{\bfseries NPMI}&\bfseries UCI\\
\hline
\multirow{10}{*}{20Newsgroups}&NVDM&-0.2558 &0.1432 &-0.0984 &-2.9496\\
&NVLDA&0.1205 &0.1763  &-0.0207 &-1.3466\\
&ProdLDA&0.1858 &0.2155 &-0.0083 &-1.5044\\
&LDA&0.2361 &0.1769  &0.0523 &0.3399 \\
&Gaussian-LDA&0.1183&0.1568&0.0252&-0.0505\\
&Scholar&0.1741& 0.1686& 0.0347&0.1497\\
&ETM&0.2437&0.1812&0.0558&0.3445\\
&ATM&0.1914&0.1720&0.0207&-0.3871\\
&VaGTM&0.3297 &0.2190& 0.0744&0.5153\\
&VaGTM-IP&{\bfseries 0.3545} &{\bfseries 0.2273} &{\bfseries 0.0843} &{\bfseries 0.6334}\\
\hline
\multirow{10}{*}{Grolier}&NVDM&-0.1877 &0.1456 &-0.0619 &-2.1149\\
&NVLDA&-0.2205 &0.1504  &-0.0653 &-2.4797\\
&ProdLDA&-0.0374 &0.1733 &-0.0193 &-1.6398\\
&LDA&0.1908 &0.2009  &0.0497&-0.0503 \\
&Gaussian-LDA&-0.0383&0.1860&-0.0115&-1.0623\\
&Scholar& 0.1388&0.1713&0.0341& -0.3144\\
&ETM& 0.1985& 0.1909&0.0541&-0.0131\\
&ATM&0.2105&0.2188&0.0582&0.1051\\
&VaGTM& 0.2515& 0.2256& 0.0706&0.2464\\
&VaGTM-IP&{\bfseries 0.2810}&{\bfseries 0.2286} &{\bfseries 0.0778} &{\bfseries 0.3683} \\
\hline
\multirow{10}{*}{NYtimes}&NVDM&-0.4130 &0.1341  &-0.1437 &-4.3072 \\
&NVLDA&-0.1575 &0.1482  &-0.0614 &-2.4208 \\
&ProdLDA&-0.0034 &0.1963  &-0.0282 &-1.9173\\
&LDA&0.3083 &0.2127 & 0.0772 &0.5165 \\
&Gaussian-LDA&-0.0707&0.1687&-0.0135&-0.9364\\
&Scholar&0.2893&0.2222&0.0635& 0.1395\\
&ETM&0.2368&0.1894&0.0603&0.2012\\
&ATM&0.3568&0.2375&0.2375&0.6582\\
&VaGTM& 0.3715& 0.2402& 0.0950&0.6718\\
&VaGTM-IP&{\bfseries 0.4304} &{\bfseries 0.2467} &{\bfseries 0.1084} &{\bfseries 0.9288}\\
\hline
\end{tabular}}
\caption{Average topic coherence on 20Newsgroups, Grolier and NYtimes with five topic settings [20, 30, 50, 75, 100].}
\label{tbs:average_coherence}
\end{table}

\begin{table*}[ht]
\centering
\small
\scalebox{0.85}{
\begin{tabular}{c|c|l}
\hline
{\bfseries Model}& \textbf{Topics}&\multicolumn{1}{c}{\bfseries Topics}\\
\hline
\multirow{4}{*}{VaGTM-IP}&Law& law bill federal legislation issue gun states protection court government\\
&Music& music song album dance band artist musical singer concert recording \\
&Film&film movie character actor show movies play star starring love \\
&War&war military soldier attack killed troop rebel commander army forces\\
\hline
\multirow{4}{*}{VaGTM}&Law&law bill federal legislation rules legal gun government proposal decision  \\
&Music &music song band sound album pop recording concert dance rock \\
&Film &film movie character actor movies love comedy drama show humor \\
&War&military war palestinian forces attack soldier army peace israeli troop \\
\hline
\multirow{4}{*}{LDA} & Law&bill law \emph{group} issue \emph{member} federal right legislation support rules \\
&Music &music song band sound record artist album show musical rock\\
&Film &film movie character play actor director movies \emph{minutes} theater cast\\
&War &war palestinian peace military soldier israeli troop attack border leader \\
\hline
\multirow{4}{*}{ProdLDA}& Law&\emph{everglades} veto negotiator \emph{billion} legislative treaty lawmaker \emph{appropriation amendment} proposal\\
&Music &musical album \emph{playwright} composer \emph{choreographer} \emph{onstage} songwriter song guitarist repertory\\
&Film &film comedy \emph{beginitalic} \emph{enditalic} sci filmmaker cinematic filmmaking movie starring\\
&War &peacekeeping military commander \emph{surplus debates trillion warhead civilian} troop \emph{interceptor}\\
\hline
\end{tabular}
}
\label{table:stu}
\caption{Topic examples extracted by selected models, italics means out-of-topic words.} 
\label{tbs:example_topics}
\end{table*}

\begin{figure*}[!h]
\centering
\includegraphics[
width=0.85\textwidth,
  keepaspectratio
  ]
{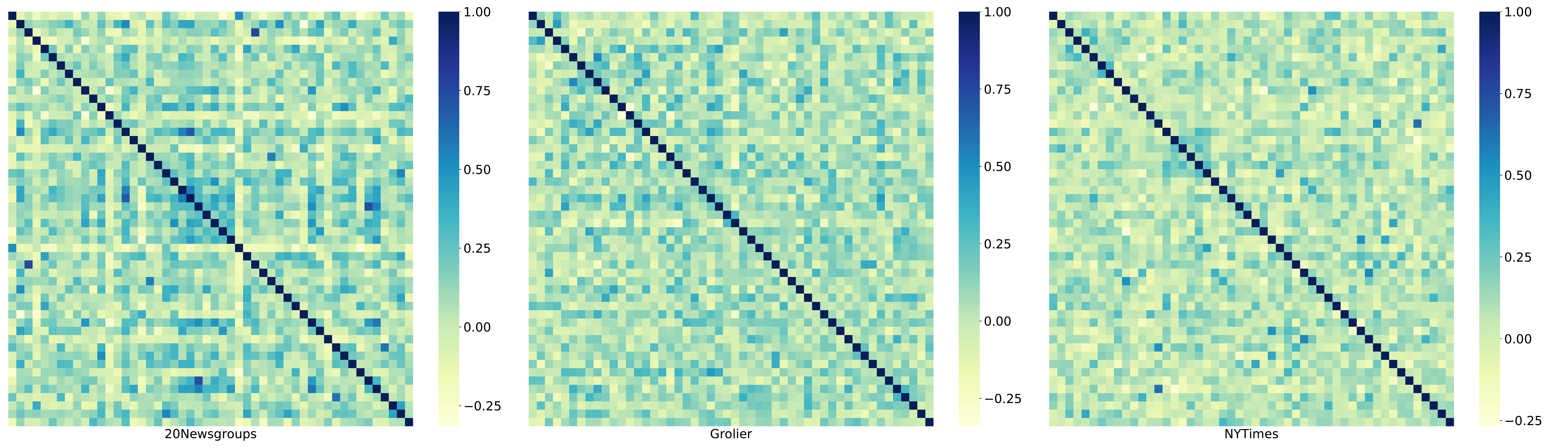}
\caption{{\color{black}The topic correlation matrix on three dataset learned by VaGTM-IP.}}
\label{heatmap}
\end{figure*}

To compare the topic extraction performance comprehensively, we firstly make a comparison of topic coherence vs. different topic proportions. In this part, we conduct the experiments on all datasets with five topic number settings [20, 30, 50, 75, 100]. Concretely, we calculate the average topic coherence among topics whose coherence are ranked at the top 70\%, 80\%, 90\% and 100\% positions. For example, to calculate the average NPMI value of VaGTM @80\%, we should first compute the average NPMI coherence with the selected topics whose NPMI values are ranked at the top 80\% for each topic number settings and then average the five averaged coherence values. The detailed comparison is shown in Fig.~\ref{fig:cmp_bar}.   

{\color{black}As the statistics shown in Fig.~\ref{fig:cmp_bar}, both VaGTM and VaGTM-IP perform competitively compared with the baseline approaches. More detailed, VaGTM-IP achieves the highest values on four metrics (C\_P, C\_A, NPMI and UCI) among all topic proportions and datasets, and VaGTM outperforms the compared baselines as well except on C\_A metric. For the 20Newsgroups dataset on C\_A, though ProdLDA performs slightly better than VaGTM (70\%, 80\% and 90\%), VaGTM obtains much higher topic coherence values considering C\_P, NPMI and UCI metrics. Thus, it would be reasonable to conclude that both VaGTM and VaGTM-IP could generally generate more coherent topics than compared approaches. And the improved performance of VaGTM and VaGTM-IP may attribute to:
 1). Both models incorporate the word relatedness into the modeling process which is helpful for grouping semantically related words into the same topic; 2) The invertible neural projector transforms word embeddings into a new embedding space and the transformed representation is more suitable for topic modeling.}

Moreover, to explore how the performance varies with different topic numbers, we further compare the average topic coherence (considering all the extracted topics) vs. different topic number settings for each dataset. Fig.~\ref{cmpcurve} depicts the detailed comparison. From the curves in the subplots, it is clear that the proposed approaches (VaGTM and VaGTM-IP) perform more stable than compared baselines.
Also, in most cases, they obtain higher average coherence values. Similar to Fig.~\ref{fig:cmp_bar}, ProdLDA performs well on C\_A metric, and it gives slightly better results than VaGTM in 20 topic settings. But beyond that, VaGTM-IP and VaGTM largely outperforms all the other baselines on all other settings. This might attribute to the incorporation of the word relatedness contained in word embeddings. We also provide a numerical comparison of average coherence values in Table~\ref{tbs:average_coherence}, each value is calculated by averaging the average topic coherences (considering all topics) over five topic number settings.

{\color{black}}From the above topic coherence comparison (Fig.~\ref{fig:cmp_bar}, Fig.~\ref{cmpcurve} and Table~\ref{tbs:average_coherence}), it is obvious that VaGTM and VaGTM-IP perform better than baselines. To validate this qualitatively, we provide four topic examples in Table~\ref{tbs:example_topics} and out-of-topic words are highlighted in italic. 

{\color{black}Besides, thanks to the usage of Gaussian distribution for topic modeling, the proposed VaGTM-IP could capture the semantic correlation between extracted topics. In detail, mean vector of learned topic-associated Gaussian could be viewed as topic embeddings in mixed Gaussian embedding space and hence the semantic relations between topics can be captured by the cosine similarity between mean vectors of topic associated Gaussian.  And Fig.~\ref{heatmap} show the visualization of topic correlation matrix for 20Newsgroups, Grolier and NYTimes datasets on 50 topic setting. Higher value implies that corresponding topics have higher semantic similarity. }

\section{Conclusion}
In this paper, we have explored the variational neural topic models and proposed Variational Gaussian Topic Model (VaGTM) and Variational Gaussian Topic Model with Invertible neural Projections (VaGTM-IP). VaGTM incorporates the word relatedness stored in pre-trained word embeddings into the modeling process to enhance the quality of extracted topics. Moreover, {\color{black}to address the issue that pre-trained word embedding is not suitable enough for specific neural topic modeling}, we extended VaGTM and proposed VaGTM-IP. It uses an invertible neural projector to transform the pre-trained word embedding into {\color{black}a mixed Gaussian embedding space which is more suitable for topic modeling}. The experimental comparison with the state-of-the-art baselines on three benchmark datasets shows that VaGTM and VaGTM-IP achieve improved topic coherence results.

\bibliography{emnlp2020}
\bibliographystyle{acl_natbib}

\end{document}